\documentclass[11pt]{article}

\usepackage[margin=1in]{geometry}
\usepackage[utf8]{inputenc}
\usepackage[T1]{fontenc}
\usepackage{lmodern}
\usepackage{microtype}
\usepackage{amsmath,amssymb,amsfonts}
\usepackage{graphicx}
\usepackage{booktabs}
\usepackage{caption}
\usepackage{subcaption}
\usepackage{titlesec}
\usepackage[round,authoryear]{natbib}
\usepackage{xcolor}
\usepackage[colorlinks=true,linkcolor=black,citecolor=blue!55!black,urlcolor=blue!55!black]{hyperref}

\titleformat{\section}{\large\bfseries}{\thesection}{0.6em}{}
\titleformat{\subsection}{\normalsize\bfseries}{\thesubsection}{0.6em}{}
\newcommand{\Cp}{C_p}
\newcommand{\Ct}{C_t}

\title{\bfseries Physics-Informed Neural Networks for Chemotherapy Pharmacokinetics:\\[2pt] Benchmarking the Clinical Estimator and Exposing Parameter Identifiability}

\author{%
  Riya Bisht\thanks{\texttt{manasi.riya2003@gmail.com}}%
  \qquad
  Dhruv Agarwal\thanks{\texttt{dhruvagarwal5018@gmail.com}}%
}
\date{May 31, 2026}

\begin{document}
\maketitle

\begin{abstract}
Physics-Informed Neural Networks (PINNs) are an attractive tool for \emph{partial-observation} problems in biology, where the governing dynamics are known but some compartments cannot be measured. Chemotherapy pharmacokinetics (PK) is a clean instance: drug concentration in plasma is routinely measured, but concentration in tissue---which determines tumour kill and off-target toxicity---is not. We benchmark a PINN against the standard clinical baseline (nonlinear least-squares on the analytical biexponential plasma solution, hereafter NLS) and a physics-agnostic neural baseline (a data-only MLP) on two PK problems. On the \textbf{linear two-compartment} problem, NLS is near-optimal; the PINN matches it to within a small constant factor while also producing the tissue curve in a single training pass, whereas the data-only MLP fails on tissue by roughly $10\times$. On a \textbf{Michaelis--Menten} extension (saturable elimination), the biexponential closed form no longer exists, so NLS is mis-specified and silently returns meaningless rate constants. The PINN instead exposes a deeper fact: the Michaelis--Menten two-compartment model is \emph{non-identifiable from plasma alone}, and the PINN reports this honestly by converging to a basin with $k_{12} \to 0$. Adding \textbf{two sparse tissue observations} largely resolves identifiability: across five seeds the PINN recovers $k_{21}$ to within 1\% of truth and $V_{\max}, K_m$ to within one standard-deviation bar, while $k_{12}$ moves in the correct direction ($0.02 \to 0.82$) but remains ${\sim}2\sigma$ below truth---a recovery the closed-form NLS estimator cannot attempt at all, because its biexponential ansatz describes only plasma. Our claim is \emph{not} that PINNs beat NLS. It is that PINNs offer a uniform recipe that ties the textbook estimator on the textbook problem, exposes structural identifiability that the textbook estimator hides, and absorbs heterogeneous measurements within a single loss.
\end{abstract}

\section{Introduction}
Chemotherapy dosing is fundamentally a question of \emph{concentration over time, at the right place}. Plasma concentration is observable through blood sampling; tissue (and tumour) concentration is generally not observable in patients. Compartmental PK models therefore play a double role: (i) summarising plasma data and (ii) inferring the unobserved compartments where the drug actually acts.

The standard estimator in clinical PK is nonlinear least-squares on a biexponential plasma function, derived analytically from the linear two-compartment ODE. It works extremely well when its assumptions hold (linear kinetics, well-characterised input, modest noise) and poorly when they do not (saturable enzymes, time-varying input, structural extensions).

PINNs \citep{raissi2019} offer an alternative formulation: rather than solve the ODE symbolically and fit its closed form, we train a neural network to satisfy the ODE in the sense of a \emph{residual}, evaluated by automatic differentiation at collocation points. Three things become easy: nonlinear and structural extensions are a one-line change to the residual; hidden compartments are returned by the same network that fits the observed compartment; and the known initial condition (the dose) is enforced as a loss term rather than a fitted parameter. This paper sets up the fairest possible comparison between the two approaches and reports the numbers honestly. We do not propose a new architecture, loss, or optimiser; our contribution is a careful, multi-seed benchmark and one identifiability observation we believe is usefully framed.

\section{Related work}
PINNs were introduced by \citet{raissi2019} for forward and inverse problems in differential equations, building on earlier neural ODE/PDE solvers \citep{lagaris1998}; \citet{karniadakis2021} and \citet{cuomo2022} survey the now-large physics-informed and scientific machine-learning literature, and mature software such as DeepXDE \citep{lu2021} has lowered the cost of applying the recipe. A central appeal for our setting is inverse inference of \emph{unobserved} fields: \citet{raissi2020} recover hidden velocity and pressure from partial flow observations, and \citet{yazdani2020} infer unobserved species and unknown parameters in biological reaction networks from sparse, noisy measurements---the systems-biology analogue of recovering an unmeasured tissue compartment from plasma. PINN training is, however, known to be delicate---loss landscapes admit poor basins and stiff-gradient pathologies \citep{wang2022}---which is precisely the lens through which we read the $k_{12} \to 0$ solution below: not an optimiser failure, but the basin the data actually support.

Applications of PINNs to compartmental pharmacokinetics are now a rapidly growing area. Recent work embeds compartmental ODEs---including fractional-order and time-varying-rate extensions---in the PINN residual to discover drug dynamics \citep{ahmadi2025cminns, ahmadi2024fractional, nasim2024}, learns unknown reaction terms in chemotherapy PK/PD models such as doxorubicin cell-kill via universal PINNs \citep{podina2024}, estimates population-level parameter \emph{distributions} from aggregated concentration data with a distributional PINN benchmarked against MCMC \citep{tsiros2026}, and performs inverse parameter estimation in physiologically-based (PBPK) models benchmarked against classical estimators \citep{wickramasinghe2025pbpk}. Relative to this line, our setting is deliberately minimal---a two-compartment model with a closed-form plasma solution---which lets us run a comparison that is still uncommon: a like-for-like, multi-seed benchmark of a \emph{standard} PINN against the nonlinear least-squares biexponential estimator that clinical pharmacologists actually use \citep{gabrielsson2016}, rather than against a data-only neural network, an MCMC sampler, or an evolutionary optimiser alone. The closest existing comparison pits scientific-ML against population-PK and classical ML for concentration \emph{prediction} \citep{valderrama2025}; our focus is instead single-patient parameter identifiability and hidden-compartment recovery.

The structural and practical non-identifiability of (bio)compartmental ODE models from partial observations is itself long established \citep{bellman1970, cobelli1980, walter1997}, including for nonlinear models in biology \citep{miao2011} and for nonlinear (Michaelis--Menten) PK models specifically \citep{godfrey1994}; the structural-versus-practical distinction is formalised by profile-likelihood analysis \citep{raue2009}, and identifiability analysis has recently been coupled directly with PINN-style parameter estimation and gray-box identification in systems biology \citep{daneker2023, aiaristotle2024}. Our contribution is not this fact but the qualitative observation that, on the Michaelis--Menten extension, a PINN \emph{surfaces} non-identifiability directly---by converging to a $k_{12} \to 0$ basin---whereas the mis-specified closed-form estimator conceals it behind rate constants that do not apply to the true ODE. This differs in stance from methods that establish identifiability \emph{a priori} and then constrain the PINN loss or model structure accordingly \citep{ahmadi2025cminns, daneker2023}: we leave the fit unconstrained and read its collapse to the $k_{12} \to 0$ basin as the diagnostic itself. To our knowledge this framing, together with the like-for-like multi-seed comparison against the clinical estimator, has not been reported.

\section{Background}
\subsection{The linear two-compartment model}
After an instantaneous intravenous bolus of dose $D$ distributed over plasma volume $V_p$, the linear two-compartment ODE is
\begin{align}
\frac{d\Cp}{dt} &= -(k_e + k_{12})\,\Cp + k_{21}\,\Ct, \\
\frac{d\Ct}{dt} &= k_{12}\,\Cp - k_{21}\,\Ct, \qquad \Cp(0) = D/V_p,\quad \Ct(0) = 0,
\end{align}
with closed-form plasma solution $\Cp(t) = A e^{-\alpha t} + B e^{-\beta t}$, where $\alpha + \beta = k_e + k_{12} + k_{21}$, $\alpha\beta = k_e k_{21}$, and $A + B = \Cp(0)$. The classical estimator fits $(A, B, \alpha, \beta)$ to plasma samples and inverts to $(k_e, k_{12}, k_{21})$.

\subsection{The Michaelis--Menten extension}
When the eliminating enzyme saturates,
\begin{equation}
\frac{d\Cp}{dt} = -\frac{V_{\max}\Cp}{K_m + \Cp} - k_{12}\Cp + k_{21}\Ct.
\end{equation}
There is no biexponential closed form. The plasma curve decays roughly linearly in time when $\Cp \gg K_m$ (zero-order, capacity-limited elimination) and roughly exponentially when $\Cp \ll K_m$ (first-order). This regime change is the hallmark of saturable (capacity-limited) elimination, the textbook clinical example being phenytoin, whose clearance falls as concentration rises within the therapeutic range.

\subsection{PINN formulation}
The network $f_\theta : \mathbb{R} \to \mathbb{R}^2$ maps time to $(\hat\Cp, \hat\Ct)$. With ODE parameters $\theta_{\mathrm{phys}}$ jointly trainable (stored as $\log\theta_{\mathrm{phys}}$ to stay positive) and collocation points $\{t_c\}$ sampled from $[0, T]$, the loss is
\begin{equation}
L = L_{\mathrm{data}} + \lambda_{\mathrm{phys}} L_{\mathrm{phys}} + \lambda_{\mathrm{ic}} L_{\mathrm{ic}},
\end{equation}
with $L_{\mathrm{data}} = \mathrm{mean}_{t\in\mathrm{obs}}(\hat\Cp(t) - \Cp^{\mathrm{obs}}(t))^2$, $L_{\mathrm{ic}} = (\hat\Cp(0) - \Cp(0))^2 + \hat\Ct(0)^2$, and $L_{\mathrm{phys}}$ the mean-squared ODE residual for both states, with $d\hat C/dt$ from \texttt{autograd}. Switching between the linear and Michaelis--Menten residual is a two-line edit.

\section{Method}
\textbf{Architecture.} A small fully-connected network suffices: 4 hidden layers $\times$ 64 units, $\tanh$ activations, and a \texttt{softplus} output to enforce non-negative concentrations (${\sim}13$k parameters).

\textbf{Losses and weights.} $\lambda_{\mathrm{phys}} = 5$, $\lambda_{\mathrm{ic}} = 20$. The IC weight is high because the dose is known exactly; the physics weight is set so the physics and data terms are comparable in magnitude after a few hundred Adam steps. Results are insensitive to $\lambda_{\mathrm{phys}} \in [1, 20]$.

\textbf{Two-stage optimisation.} Standard PINN practice: Adam \citep{kingma2015} for global exploration (6000--8000 steps, lr $5\times10^{-3}$, 200 collocation points resampled each step---70\% uniform on $[0, 24]$, 30\% concentrated on $[0, 4]$ where the dynamics are fast), then L-BFGS \citep{liu1989} with strong-Wolfe line search (up to ${\sim}300$ iterations on a fixed 160-point grid). Wall-clock per PINN run is ${\sim}20$--$30$s on a single CPU core; the whole benchmark finishes in under an hour on a laptop.

\textbf{Baselines. NLS:} \texttt{scipy.optimize.curve\_fit} (Levenberg--Marquardt) on $\Cp(t) = (\Cp{}_0 - B)e^{-\alpha t} + B e^{-\beta t}$, fixing $A + B = \Cp{}_0$ (known dose), then inverting $(A, B, \alpha, \beta) \to (k_e, k_{12}, k_{21})$ and integrating the ODE for the tissue curve. \textbf{MLP:} identical network and optimiser to the PINN, plasma data only, no physics term and no tissue supervision---this isolates the contribution of the physics prior.

\textbf{Data.} Synthetic ground truth from \texttt{scipy.integrate.odeint} on a 481-point grid over $[0, 24]$ hours. Sparse plasma samples are drawn at evenly spaced times in $[0.5, 24]$\,h and corrupted with multiplicative Gaussian noise ($\Cp^{\mathrm{obs}} = \Cp^{\mathrm{true}}(1 + \varepsilon)$, $\varepsilon \sim \mathcal{N}(0, \sigma^2)$, clipped to $\ge 0$). Reference parameters are in the published range for doxorubicin ($k_e = 0.5$, $k_{12} = 1.0$, $k_{21} = 0.4$, $V_p = 25$\,L, $D = 50$\,mg).

\section{Experiments}
We run three linear-ODE experiments at five seeds each (mean $\pm$ std reported throughout): a \emph{headline} run ($n = 8$, $\sigma = 10\%$); a \emph{noise sweep} ($\sigma \in \{1, 5, 10, 20\}\%$, $n = 8$); and a \emph{sparsity sweep} ($n \in \{4, 6, 8, 12, 20\}$, $\sigma = 10\%$). We then run a structural-misspecification experiment in which data is generated by the Michaelis--Menten ODE and all three methods are applied. All raw runs are released with the code.

\section{Results}
\subsection{Headline: linear ODE}
Table~\ref{tab:headline} should be read honestly: NLS and PINN are statistically tied on every metric (differences within one std bar across five seeds). NLS recovers $k_e$ both more cleanly and with less bias---the PINN is marginally biased high (0.520 vs.\ true 0.500)---because it is the maximum-likelihood estimator on a correctly-specified model; the PINN pays a ${\sim}6000\times$ wall-clock penalty for the otherwise-equivalent answer. The physics-agnostic MLP, with identical capacity, is ${\sim}10\times$ worse on the hidden tissue compartment and ${\sim}4\times$ worse on plasma---that gap is the clean PINN-vs-no-prior demonstration. The large std on $k_{12}$ for both estimators is a \emph{practical} (conditioning) effect, not a structural one: given the known dose the linear two-compartment model is structurally identifiable from plasma alone---in the noise-free limit the biexponential inversion $k_{21} = (A\beta + B\alpha)/(A+B)$, $k_e = \alpha\beta/k_{21}$, $k_{12} = \alpha + \beta - k_e - k_{21}$ recovers $(k_e, k_{12}, k_{21})$ exactly---but $k_{12}$ is by far the most poorly conditioned of the three at realistic noise, because it is obtained only as a difference of comparably-sized quantities \citep{raue2009}. This is a property of the data, not a flaw of either method.

\begin{table}[t]
\centering
\caption{Headline linear-ODE result ($n = 8$ samples, $\sigma = 10\%$, 5 seeds). Best value per column in bold. RMSE in mg/L.}
\label{tab:headline}
\begin{tabular}{lcccccc}
\toprule
Method & RMSE $\Cp$ & RMSE $\Ct$ & $k_e\ (0.50)$ & $k_{12}\ (1.00)$ & $k_{21}\ (0.40)$ & Wall-clock \\
\midrule
NLS & $\mathbf{0.016 \pm 0.017}$ & $\mathbf{0.031 \pm 0.025}$ & $\mathbf{0.493 \pm 0.013}$ & $0.921 \pm 0.215$ & $0.368 \pm 0.090$ & 0.004\,s \\
PINN & $0.017 \pm 0.017$ & $0.039 \pm 0.031$ & $0.520 \pm 0.018$ & $0.883 \pm 0.208$ & $0.361 \pm 0.064$ & 25.1\,s \\
MLP & $0.071 \pm 0.020$ & $0.355 \pm 0.111$ & --- & --- & --- & 9.2\,s \\
\bottomrule
\end{tabular}
\end{table}

\subsection{Noise and sparsity robustness}
Figure~\ref{fig:robust} summarises tissue-RMSE behaviour across noise levels and sample counts (full tables in Appendix~\ref{app:robust}). NLS scales tightly and essentially linearly with noise, as expected for a correctly-specified linear estimator. The PINN scales similarly with a slightly higher floor at low noise (the physics residual caps how aggressively it fits individual plasma points); at realistic clinical noise ($\ge 10\%$) the two are within one std bar. The PINN is within ${\sim}1.5\times$ of NLS for $n \ge 8$ and degrades at $n = 4$, where the network has more capacity than the data can constrain. The MLP tissue error is roughly flat regardless of noise or sample count---there is no information about tissue in plasma samples without a physics prior to propagate it.

\begin{figure}[t]
\centering
\begin{subfigure}{0.49\linewidth}\includegraphics[width=\linewidth]{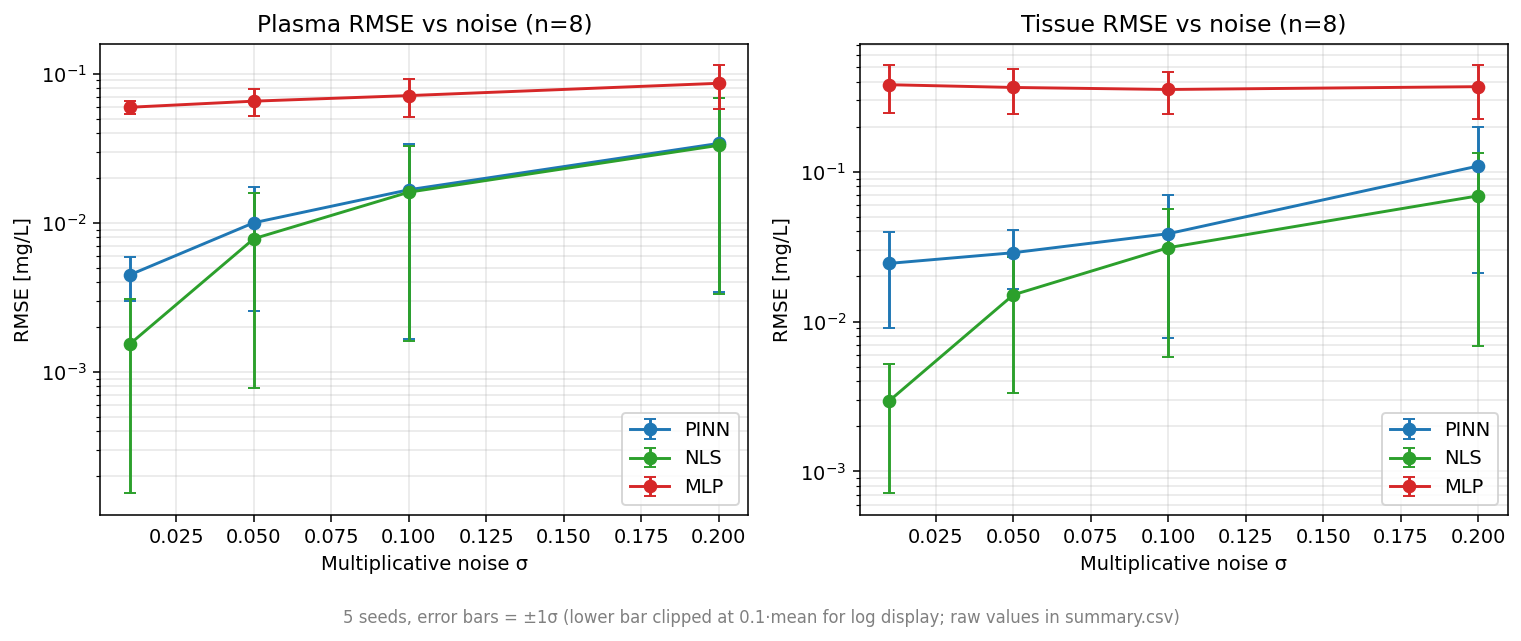}\end{subfigure}
\hfill
\begin{subfigure}{0.49\linewidth}\includegraphics[width=\linewidth]{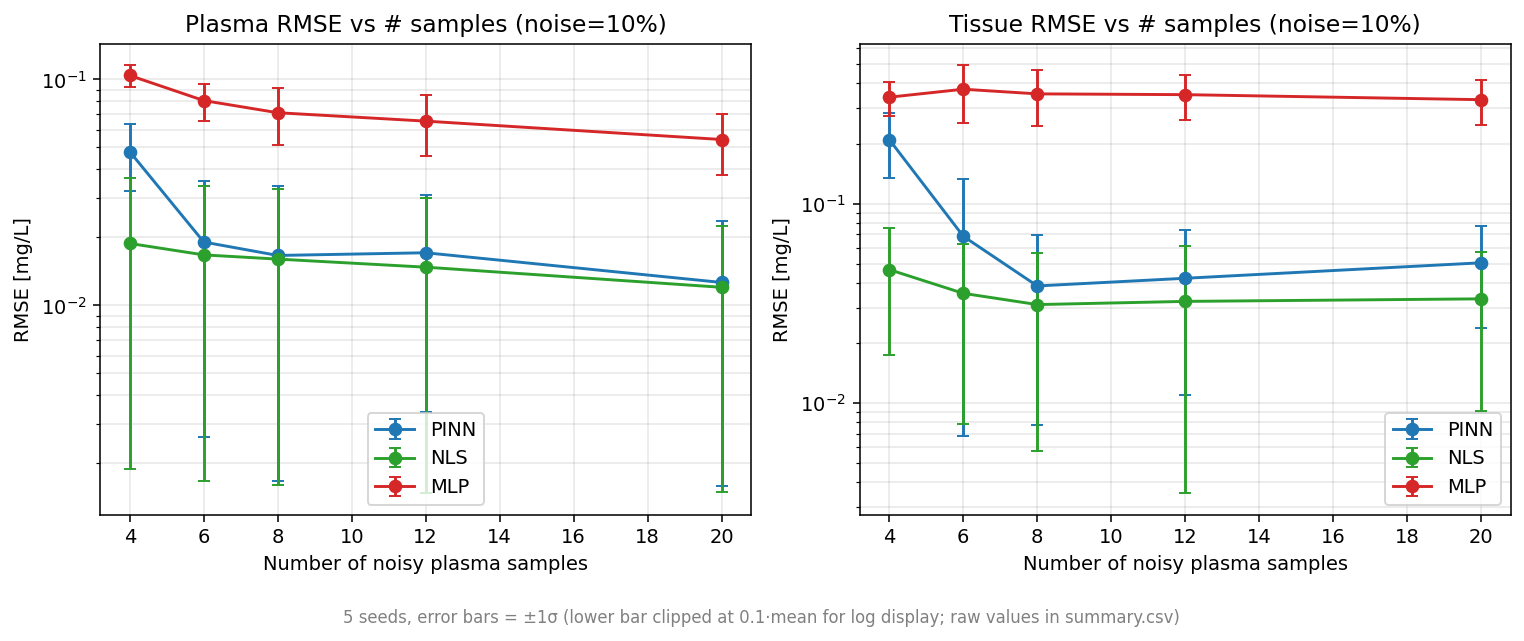}\end{subfigure}
\caption{Tissue RMSE (mg/L, log scale, error bars over 5 seeds) versus noise level (left) and plasma sample count (right). NLS and PINN track each other on the well-specified linear problem; the data-only MLP is flat and far above both.}
\label{fig:robust}
\end{figure}

\subsection{Misspecification and identifiability (Michaelis--Menten)}
This experiment exposes the \emph{structural} difference between the methods. Data is generated from the Michaelis--Menten ODE ($V_{\max} = 1.5$ mg/L/hr, $K_m = 0.3$ mg/L, $k_{12} = 1.0$, $k_{21} = 0.4$).

\textbf{Finding 1 --- NLS is mis-specified.} The biexponential closed form NLS fits does not exist for a Michaelis--Menten ODE. NLS returns the best biexponential approximation to plasma; the inverted ``linear'' rate constants are meaningless and the propagated tissue curve has the wrong shape (peak ${\sim}3\times$ too low).

\textbf{Finding 2 --- the PINN with plasma alone is non-identifiable.} The PINN matches the plasma curve well (RMSE ${\sim}26\,\mu$g/L) but does so by setting $k_{12} \approx 0$ and inflating $V_{\max}$, yielding an essentially zero tissue curve. This is wrong, but it is \emph{consistent with the plasma observations}: with $k_{12} = 0$ no drug reaches tissue and plasma must clear through $V_{\max}$ alone. With four free parameters and one observable compartment, the system is locally non-identifiable in this regime: as $k_{12} \to 0$ the tissue compartment decouples from plasma, so plasma data carry essentially no information about $k_{12}$ and the Fisher information matrix degenerates at this boundary. The PINN faithfully reports the answer the data actually supports; NLS hides the same gap behind rate constants that do not apply to the true ODE. We demonstrate this empirically rather than prove it; a formal treatment via the Fisher information matrix, profile likelihood \citep{raue2009}, or a structural-identifiability analysis of the nonlinear model \citep{cobelli1980, godfrey1994} is left to future work.

\begin{figure}[t]
\centering
\includegraphics[width=0.95\linewidth]{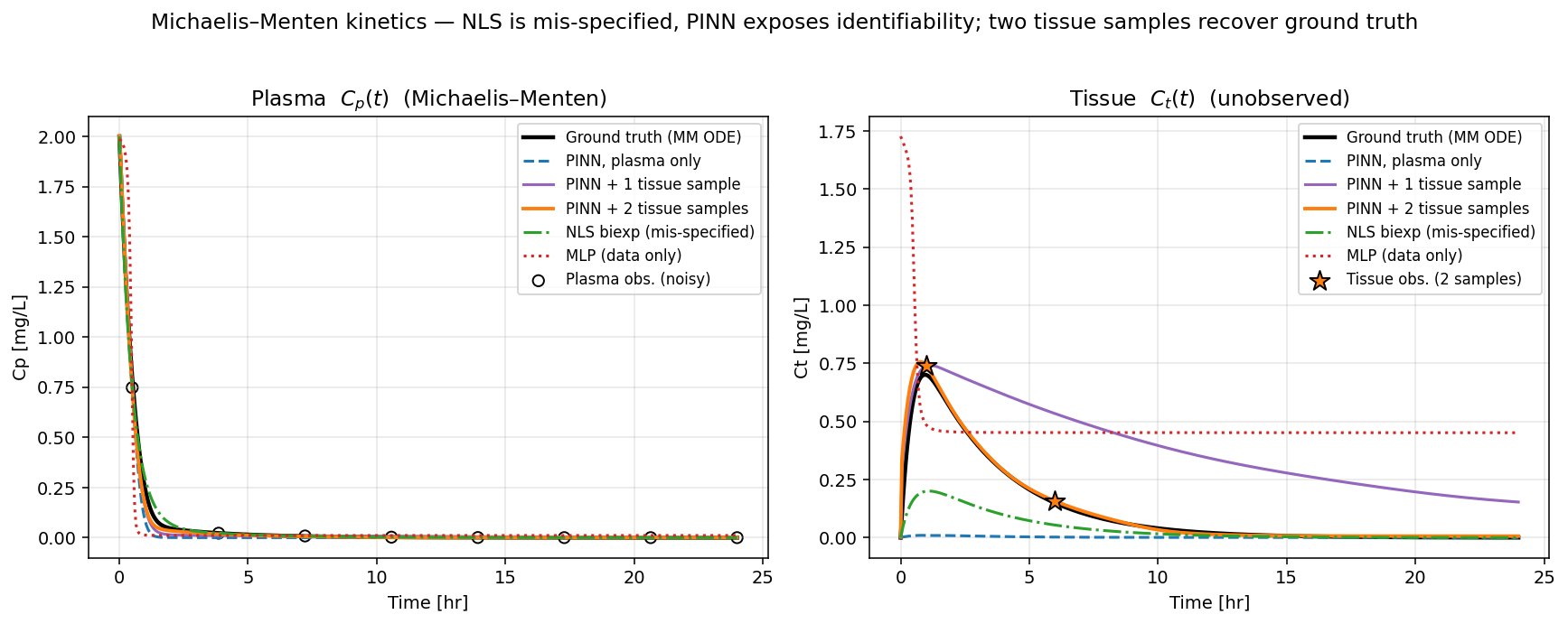}
\caption{Michaelis--Menten tissue trajectories. The ``PINN + 2 tissue'' curve overlays the ground-truth tissue trajectory almost exactly; NLS (mis-specified), the data-only MLP, and the plasma-only PINN each fail differently. Plasma-only PINN collapses to near-zero tissue ($k_{12} \to 0$), the honest report of a non-identifiable fit.}
\label{fig:mm}
\end{figure}

\textbf{Finding 3 --- sparse heterogeneous data resolves identifiability.} Adding tissue measurements to the PINN loss is a one-line change. The \emph{closed-form} NLS estimator cannot absorb a tissue measurement, because its biexponential ansatz describes only plasma. A numerical least-squares fit of the full ODE system could in principle absorb tissue data; doing so abandons the closed-form clinical estimator and amounts to a non-neural version of the same residual-fitting the PINN performs. Our point is therefore that the \emph{standard clinical} biexponential estimator cannot, whereas the PINN does so with a one-line change. As Tables~\ref{tab:mm-rmse} and \ref{tab:mm-param} show, one tissue sample (near the peak, $t = 1$\,h) fixes the peak amplitude but not the decay; two tissue samples ($t = 1$\,h and $t = 6$\,h) constrain both branches, recovering $k_{21}$ to within 1\% of truth and $V_{\max}, K_m$ to within one std bar across five seeds; $k_{12}$ is recovered in the correct direction ($0.02 \to 0.82$) but remains ${\sim}2\sigma$ below truth (Table~\ref{tab:mm-param}). Figure~\ref{fig:mm} shows the ``PINN + 2 tissue'' curve overlaying ground truth almost exactly, while every other method fails in a characteristic way.

\begin{table}[t]
\centering
\caption{Michaelis--Menten RMSE (mg/L, 5 seeds, $n = 8$ plasma observations).}
\label{tab:mm-rmse}
\begin{tabular}{lcc}
\toprule
Method & Plasma RMSE & Tissue RMSE \\
\midrule
NLS biexp (mis-specified) & $0.025 \pm 0.013$ & $0.166 \pm 0.012$ \\
MLP (data only) & $0.091 \pm 0.002$ & $0.492 \pm 0.164$ \\
PINN, plasma only & $0.026 \pm 0.003$ & $0.215 \pm 0.004$ \\
PINN + 1 tissue sample & $0.021 \pm 0.006$ & $0.330 \pm 0.072$ \\
\textbf{PINN + 2 tissue samples} & $\mathbf{0.015 \pm 0.012}$ & $\mathbf{0.028 \pm 0.001}$ \\
\bottomrule
\end{tabular}
\end{table}

\begin{table}[t]
\centering
\caption{Recovered Michaelis--Menten parameters (5 seeds). True values in parentheses.}
\label{tab:mm-param}
\begin{tabular}{lcccc}
\toprule
Method & $V_{\max}\ (1.50)$ & $K_m\ (0.30)$ & $k_{12}\ (1.00)$ & $k_{21}\ (0.40)$ \\
\midrule
PINN, plasma only & $3.16 \pm 0.18$ & $0.50 \pm 0.07$ & $0.02 \pm 0.02$ & $0.19 \pm 0.01$ \\
PINN + 1 tissue & $2.52 \pm 0.53$ & $0.64 \pm 0.06$ & $0.59 \pm 0.09$ & $0.06 \pm 0.01$ \\
\textbf{PINN + 2 tissue} & $\mathbf{1.91 \pm 0.46}$ & $\mathbf{0.37 \pm 0.10}$ & $\mathbf{0.82 \pm 0.09}$ & $\mathbf{0.40 \pm 0.05}$ \\
\bottomrule
\end{tabular}
\end{table}

The take-away is \emph{not} ``PINN beats NLS.'' It is that the PINN is a uniform framework that (a) ties the textbook estimator on the textbook problem, (b) exposes structural identifiability the textbook estimator hides, and (c) absorbs heterogeneous measurements---plasma points, one biopsy, a follow-up biopsy---within the same loss, recovering ground truth when the data are sufficient.

\section{Discussion}
\textbf{What this benchmark says, and what it does not.} PINNs are roughly competitive with the textbook PK estimator on the textbook PK problem, and meaningfully better when its assumptions break; they are dramatically better than physics-agnostic networks for inferring hidden states from partial observations. It does \emph{not} say PINNs are the right tool when a closed-form estimator exists, problems are well-specified, and compute is the binding constraint---there, NLS is.

\textbf{Identifiability.} For the linear model, the plasma decay fixes the macro constants $\alpha, \beta$ and the amplitude split fixes $k_{21}$; given the known dose ($A + B = \Cp{}_0$) the micro-constants $(k_e, k_{12}, k_{21})$ are then \emph{structurally} identifiable from plasma, the noise-free inversion being exact. What is poor is the \emph{conditioning}: $k_{12}$ is recovered as a difference of comparable quantities, so its seed-to-seed variance is large for NLS and PINN alike. The PINN inherits this conditioning faithfully.

\textbf{Cost.} A PINN run is ${\sim}25$\,s on CPU; an NLS fit is instant. For 50{,}000-patient population studies this matters; for single-patient model exploration, teaching, or prototyping an extension, the absolute time is negligible.

\textbf{Broader impact.} The work is a methodological benchmark on synthetic data with no patient involvement. A positive impact is clearer, more honest inference of unobserved tissue exposure, which is the quantity that governs efficacy and toxicity; PINNs that expose identifiability gaps could reduce over-confident clinical inferences. The risk is the converse: any model that outputs unobserved tissue concentrations could be over-trusted in a dosing decision. We therefore stress that none of these methods is validated on real clinical data and none should inform patient care in its present form.

\section{Limitations}
\begin{itemize}
\item \textbf{Single noise model.} We use multiplicative Gaussian noise. Real assay noise is often heteroscedastic and includes lower-limit-of-quantification censoring, which neither baseline handles out of the box.
\item \textbf{No inter-individual variability.} Real PK studies estimate population parameters with random effects (NONMEM, Monolix, Bayesian hierarchical models). All three methods here are individual-fit only.
\item \textbf{Synthetic ground truth.} We have not tested on real clinical PK data; external validity requires refitting on a published dataset (e.g.\ the doxorubicin studies of \citet{speth1988}).
\item \textbf{No methodological novelty.} The architecture, loss, and optimisation recipe are standard \citep{raissi2019}; the identifiability fact is long known in PK theory. Our contribution is the rigorous comparison and the framing, not a new method.
\end{itemize}

\section{Future work}
Natural extensions: (1) \textbf{PK/PD coupling}---add a tumour compartment with cell-kill proportional to $\Ct \cdot N$, so the PINN jointly identifies PK and PD parameters while NLS would need a separate stage; (2) a \textbf{spatial PINN} (Krogh cylinder) using $\partial^2 C/\partial x^2$ from autograd for radial drug diffusion; (3) a \textbf{heteroscedastic/LLOQ-aware likelihood} replacing MSE with a censored-Gaussian negative log-likelihood; (4) a \textbf{population PINN} with per-patient parameter heads on a shared backbone; and (5) a \textbf{Bayesian PINN} \citep{bpinns2021} that turns the qualitative non-identifiability signal into calibrated posterior uncertainty on $(k_{12}, V_{\max}, K_m)$.

\bibliographystyle{plainnat}

\appendix
\section{Full robustness tables}
\label{app:robust}

\begin{table}[h]
\centering
\caption{Tissue RMSE (mg/L) across multiplicative noise levels ($n = 8$, 5 seeds). Best per row in bold.}
\label{tab:noise}
\begin{tabular}{lccc}
\toprule
$\sigma$ & PINN & NLS & MLP \\
\midrule
1\% & $0.024 \pm 0.015$ & $\mathbf{0.003 \pm 0.002}$ & $0.382 \pm 0.135$ \\
5\% & $0.029 \pm 0.012$ & $\mathbf{0.015 \pm 0.012}$ & $0.366 \pm 0.123$ \\
10\% & $0.039 \pm 0.031$ & $\mathbf{0.031 \pm 0.025}$ & $0.355 \pm 0.111$ \\
20\% & $0.109 \pm 0.088$ & $\mathbf{0.069 \pm 0.064}$ & $0.370 \pm 0.145$ \\
\bottomrule
\end{tabular}
\end{table}

\begin{table}[h]
\centering
\caption{Tissue RMSE (mg/L) across plasma sample counts ($\sigma = 10\%$, 5 seeds). Best per row in bold.}
\label{tab:samples}
\begin{tabular}{lccc}
\toprule
$n$ & PINN & NLS & MLP \\
\midrule
4 & $0.209 \pm 0.074$ & $\mathbf{0.047 \pm 0.029}$ & $0.342 \pm 0.066$ \\
6 & $0.069 \pm 0.064$ & $\mathbf{0.035 \pm 0.028}$ & $0.374 \pm 0.121$ \\
8 & $0.039 \pm 0.031$ & $\mathbf{0.031 \pm 0.025}$ & $0.355 \pm 0.111$ \\
12 & $0.042 \pm 0.031$ & $\mathbf{0.032 \pm 0.029}$ & $0.352 \pm 0.089$ \\
20 & $0.050 \pm 0.027$ & $\mathbf{0.033 \pm 0.024}$ & $0.332 \pm 0.083$ \\
\bottomrule
\end{tabular}
\end{table}

\end{document}